\documentclass[twoside]{article}

\usepackage[utf8]{inputenc} 
\usepackage[T1]{fontenc}    
\usepackage{hyperref}       
\usepackage{booktabs}       
\usepackage{amsfonts}       
\usepackage{nicefrac}       
\usepackage{microtype}      
\usepackage{amsmath, amsfonts, amssymb, amstext, amsthm, bbm, mathtools}
\usepackage{physics}
\usepackage{xcolor}
\usepackage{subcaption}
\usepackage[normalem]{ulem}
\usepackage{graphicx}
\newtheorem{proposition}{Proposition}

\newtheorem*{theorem*}{Proposition}

\newcommand{\ie}{i.e.}
\usepackage{bm}
\usepackage{algorithm}
\usepackage[noend]{algorithmic}

\def\C{{\bf C}}
\def\G{{\bf G}}

\def\X{{\bf X}}
\def\Xsk{{{\bf X}^{(k)}}}
\def\xsi{{{\bf x}^{(k)}_i}} 
\def\muk{\mu^{(k)}}
\def\muck{\mu^{(k)}_c}

\def\mik{{m^{(k)}_i}}
\def\dik{\delta_\xsi}
\def\dXk{{\bf \delta}_ \Xsk}
\def\dX{{\bf \delta}_ \X}
\def\mk{{\bf m}^{(k)}}
\def\mt{{\bf m}}

\def\hck{{h^{(k)}_c}}
\def\hk{{\bf h}^{(k)}}

\def\htar{{\bf h}}

\def\Dmk{{\bf D}_1^{(k)}}
\def\Dhk{{\bf D}_2^{(k)}}

\def\ysi{y^{(k)}_i}
\def\yi{y_i}
\def\nk{{n^{(k)}}}
\def\nt{{n}}
\def\ga{{\bf \gamma}}
\newcommand{\x}{{\bf x}}

\renewcommand{\xi}{{\bf x}_i}

\newcommand{\one}{{\bf {1}}}
\newcommand*{\MyDef}{\mathrm{def}}
\newcommand*{\eqdefU}{\ensuremath{\mathop{\overset{\MyDef}{=}}}}

\newcommand{\argmin}{\mathop{\mathrm{arg\,min}}}

\newcommand{\kldiv}{\mathop{\mathrm{KL}}}
\newcommand{\diag}{\mathop{\mathrm{diag}}}

\newcommand{\nico}[1]{{#1}}
\newcommand{\rf}[1]{{#1}}

\usepackage[accepted]{aistats2019}
\usepackage[round]{natbib}

\begin{document}

%

%
\runningauthor{Ievgen Redko, Nicolas Courty, R\'emi Flamary, Devis Tuia}

\twocolumn[

\aistatstitle{Optimal Transport for Multi-source Domain Adaptation under Target Shift}

\aistatsauthor{Ievgen Redko \And Nicolas Courty}

\aistatsaddress{Univ Lyon, UJM-Saint-Etienne, CNRS\\
Institut d Optique Graduate School\\
Laboratoire Hubert Curien UMR 5516\\
F-42023, Saint-Etienne, France \And  University of Bretagne Sud, CNRS and INRIA\\
  IRISA, UMR 6074\\
  F-56000, Vannes, France} 
  
 \aistatsauthor{R\'emi Flamary \And Devis Tuia}
 \aistatsaddress{University of C\^ote d'Azur, OCA\\ Laboratoire Lagrange, CNRS UMR 7293\\
 F-06108, Nice, France \And Laboratory of Geoscience and Remote Sensing\\
  Wageningen University and Research\\
  6700, Wageningen, Netherlands}
]

\begin{abstract}
        In this paper, we tackle the problem of reducing discrepancies between multiple domains, i.e.  \emph{multi-source domain adaptation}, and consider it under the \emph{target shift} assumption: in all domains we aim to solve a classification problem with the same output classes, but with different labels proportions. This problem, generally ignored in the vast majority of domain adaptation papers, is nevertheless critical in real-world applications, and we theoretically show its impact on the success of the adaptation. Our proposed method is based on optimal transport, a theory that has been successfully used to tackle adaptation problems in machine learning. 
        The introduced approach, Joint Class Proportion and Optimal Transport (JCPOT), performs multi-source adaptation and target shift correction simultaneously by learning the class probabilities of the unlabeled target sample and the coupling allowing to align two (or more) probability distributions. Experiments on both synthetic and real-world data (satellite image pixel classification) task show the superiority of the proposed method over the state-of-the-art.
\end{abstract}

\section{INTRODUCTION}
\label{sec:intro}
In many real-world applications, it is desirable to use models trained on largely annotated data sets (or \emph{source domains}) to label a newly collected, and therefore unlabeled data set (or \emph{target domain}). However, differences in the probability distributions between them hinder the success of the direct application of learned models to the latter. To overcome this problem, recent machine learning research has devised a family of techniques, called domain adaptation (DA), that deals with situations where {source and target samples} follow different probability distributions~\citep{Qui09,Pat15}. The inequality between the joint distributions can be characterized in a variety of ways depending on the assumptions made about the conditional and marginal distributions. Among those, arguably the most studied scenario called covariate shift (or sample selection bias) considers the situation where the inequality between probability density functions (pdfs) is  due to the change in the marginal distributions ~\citep{DBLP:conf/icdm/ZadroznyLA03,Bickel:2007:DLD:1273496.1273507,nips07:Huang, DBLP:conf/nips/LiuZ14, DBLP:conf/icml/WenYG14,Fernando:2013:UVD:2586117.2587168,DBLP:conf/nips/CourtyFHR17}. 

Despite the large number of methods proposed in the literature to solve the DA problem under the covariate shift assumption, very few considered the {(widely occurring)} situation where the changes in the joint distribution is caused by a shift in the distribution of the outputs, a setting that has been referred to as \textit{target shift}. In practice, {the target shift assumption implies} that a change in the class proportions across domains is at the base of the shift: such a situation is also known as choice-based or endogenous stratified sampling in econometrics \citep{econometrica} or as prior probability shift \citep{Storkey09whentraining}.
In the classification context, target shift was first introduced in \citep{Japkowicz:2002:CIP:1293951.1293954} and referred to as the class imbalance problem. In order to solve it, several approaches were proposed. In \citep{llw-svmcns-02}, authors assumed that the shift in the target distribution was known a priori, while in~\citep{DBLP:conf/pakdd/YuZ08}, partial knowledge of the target shift was supposed to be available. In both cases, the assumption of prior knowledge about the class proportions in the target domain seems quite restrictive. 
More recent methods that avoid making this kind of assumptions are \citep{Chan:2005:WSD,conf/icml/ZhangSMW13}. In the former, authors used a variation of the {Expectation Maximization} {algorithm}, which relies on a computationally expensive estimation of the conditional distribution. In the latter, authors estimate the proportions in both domains directly from observations. Their approach, however, also relies on computationally expensive optimization over kernel embeddings of probability functions. 
\nico{This last line of work has been extended in~\citep{zhang2015multi} to the multi-domain setting, and as such constitutes a relevant baseline for our work. }{The algorithm proposed by the authors produces a set of hypotheses with one hypothesis per domain that can be combined using the theoretical study on multi-source domain adaptation presented in~\citep{mansour2009domain}.}
Despite the rather small corpus of works in the literature dealing with the subject, target shift often occurs in practice, especially in applications dealing with anomaly/novelty detection \citep{Blanchard:2010:SND:1756006.1953028,pmlr-v30-Scott13,pmlr-v33-sanderson14}, or in tasks where spatially located training sets are used to classify wider  areas, as in remote sensing image classification~\citep{tuia15,zhang2015multi}. 

\begin{figure*}[!t]
  \centering
  \begin{minipage}{0.22\linewidth}
  	\includegraphics[height=4cm]{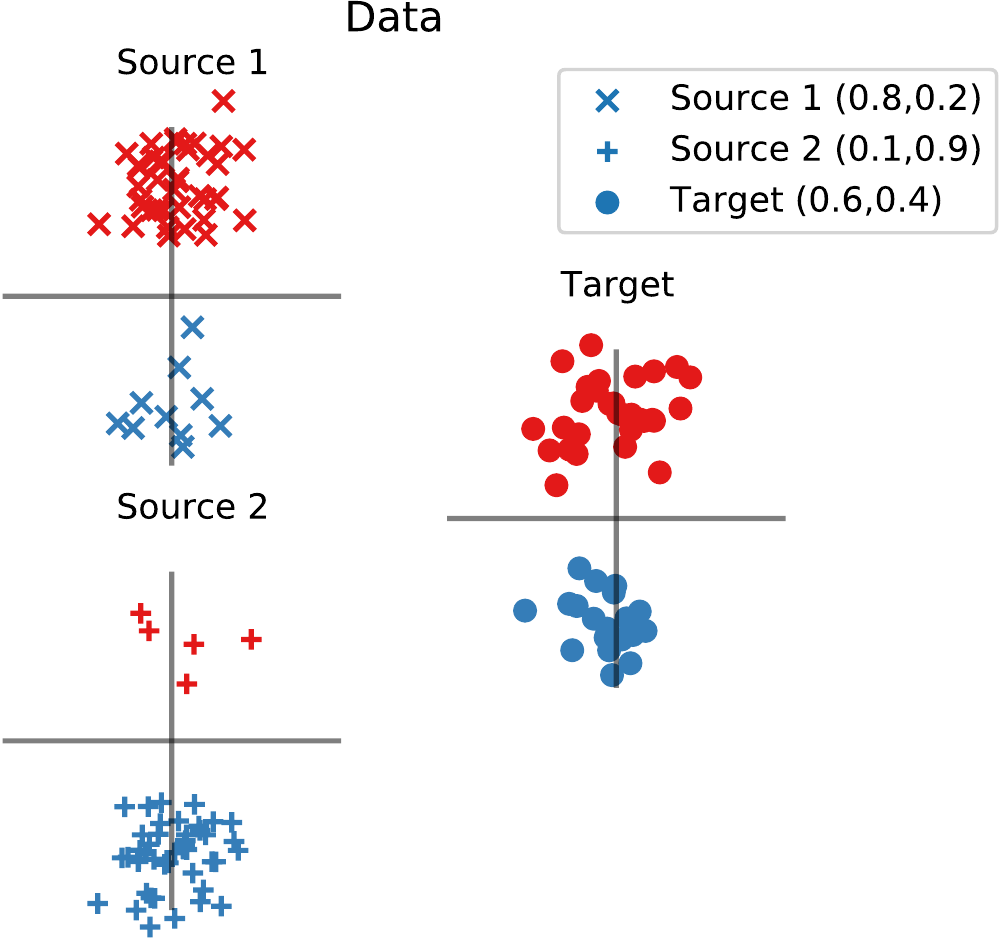}
    \subcaption{}
  \end{minipage}
  \hspace{0.3cm}
  \begin{minipage}{0.22\linewidth}
  	\includegraphics[height=4cm]{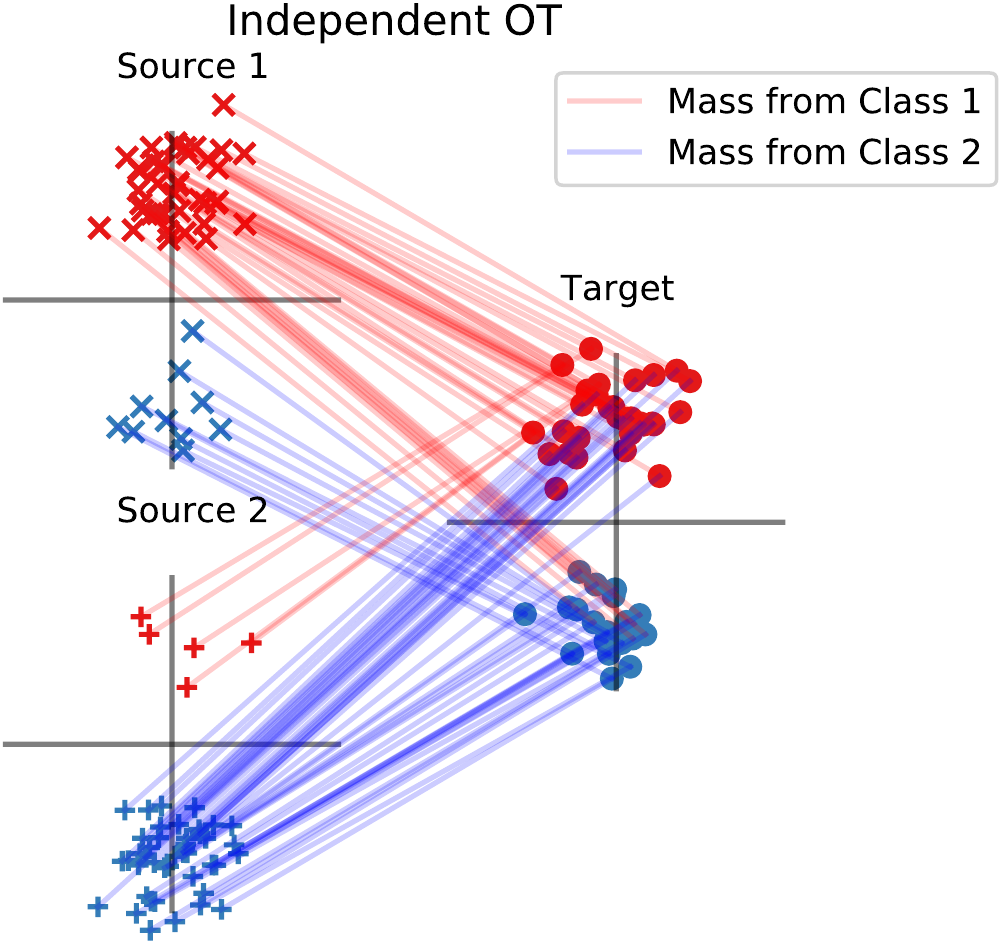}
    \subcaption{}
  \end{minipage}  
  \hspace{0.3cm}
  \begin{minipage}{0.22\linewidth}
    \includegraphics[height=4cm]{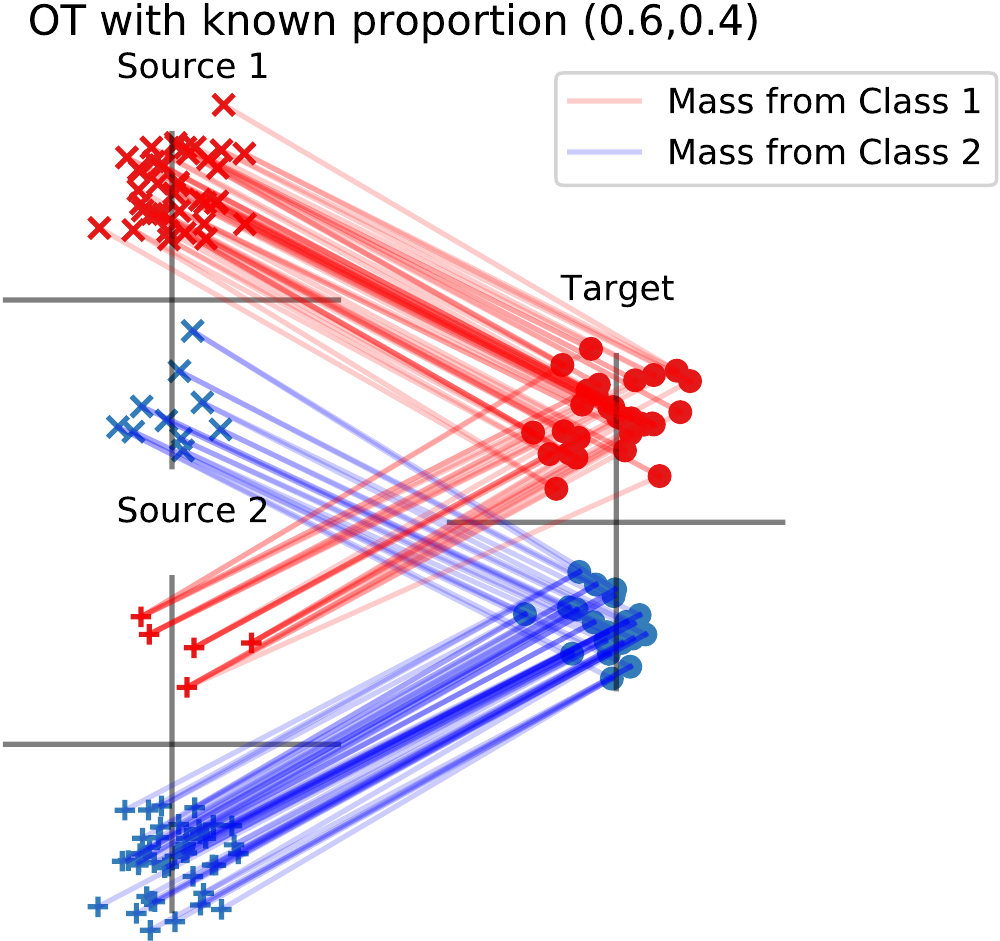}
    \subcaption{}
  \end{minipage}
  \hspace{0.3cm}
  \begin{minipage}{0.22\linewidth}
  	\includegraphics[height=4cm]{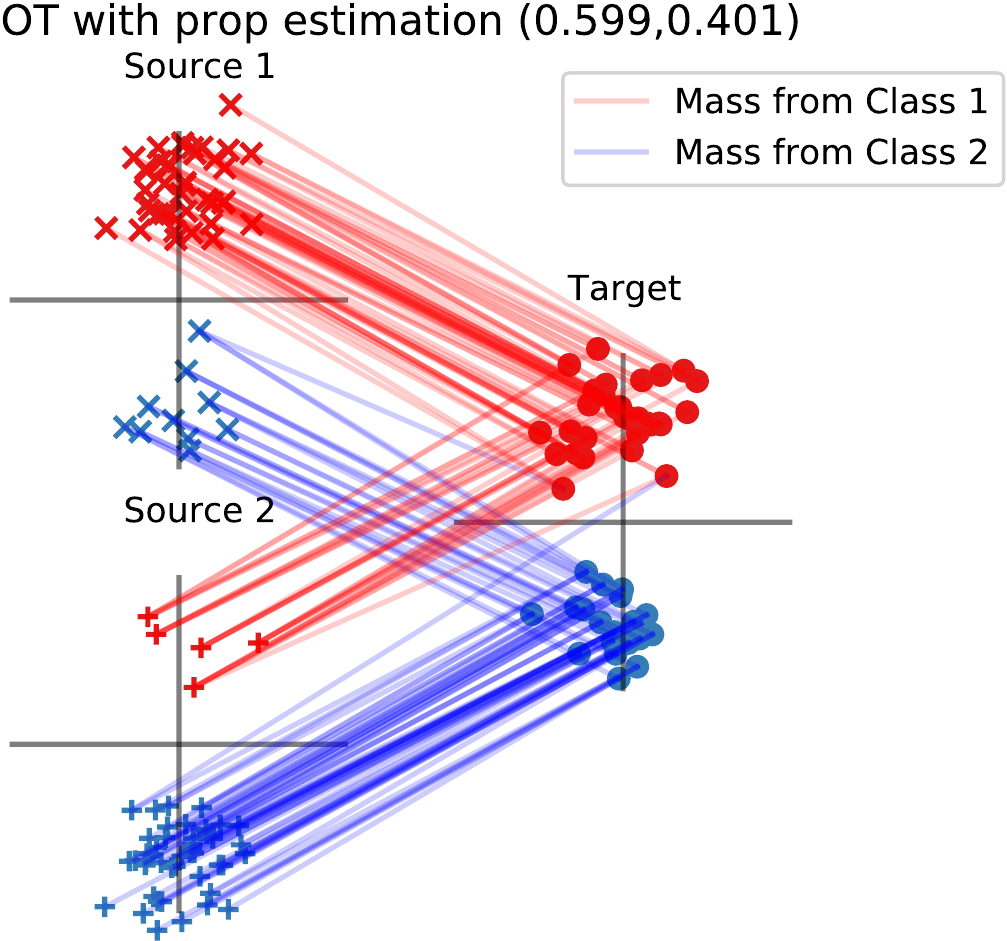}
    \subcaption{}
  \end{minipage}
  \caption{Illustration of the importance of proportion estimation for target shift: (a) the data of 2 source and 1 target domains with different class proportions is visualized; (b) DA method based on OT \citep{courty14} transports instances across different classes due to class proportions imbalance; (c) the transportation obtained when the true class proportions are used to reweigh instances; (d) transportation obtained with JCPOT that is nearly identical to the one obtained with an a priori knowledge about the class proportions. 
  }
  \label{fig:illus_ts}
\end{figure*}

In this paper, we propose a new algorithm for correcting the target shift based on optimal transport (OT). OT theory is a branch of mathematics initially introduced by Gaspard Monge for the task of resource allocation \citep{monge_81}. Originally, OT tackled a problem of aligning two probability measures in a way that minimizes the cost of moving a unit mass between them, while preserving the original marginals. 
The recent appearance of efficient formulations of OT \citep{cuturi:2013} has allowed its application in DA, as OT allows to learn explicitly the transformation of a given source pdf into the pdf of the target sample. In this work, we build upon a recent work on DA \citep{DBLP:journals/pami/CourtyFTR17}, where authors successfully casted the DA problem as an OT problem, and extend it to deal with the target shift setting. Our motivation to propose new specific algorithms for target shift stems from the fact that many popular DA algorithms designed to tackle covariate shift cannot handle the target shift equally well. This is illustrated in Figure \ref{fig:illus_ts}, where we show that the DA method based on OT mentioned above fails to restrict the transportation of mass across instances of different classes when the class proportions of source and target domains differ. However, as we show in the following sections, our \emph{Joint Class Proportion and Optimal Transport (JCPOT)} model manages to do it correctly. Furthermore and contrary to the original contribution, we also consider the much more challenging case of multi-source domain adaptation, where more than one source domains with changing distributions of outputs are used for learning. To the best of our knowledge, this is the first multi-source DA algorithm that efficiently leverages the target shift and shows an increasing performance with the increasing number of source domains considered.

The rest of the paper is organized as follows: in Section~\ref{sec:OT}, we  present regularized OT and its application to DA. Section~\ref{sec:limits} details the target shift problem and provides a generalization bound for this learning scenario \rf{and a proof that minimizing the Wasserstein distance between two distributions with class imbalance leads to the optimal solution}. In Section~\ref{sec:JCPOT}, we present the proposed JCPOT method for unsupervised DA when no labels are used for adaptation.
In Section~\ref{sec:experiments}, we provide comparisons to state-of-art methods for synthetic data in the multi-source adaptation scenario and we report results for a real life case study performed in remote sensing pixel classification.

\section{OPTIMAL TRANSPORT}\label{sec:OT}
In this section we introduce key concepts of optimal transport and some important results used in the following sections.  
\subsection{Basics and Notations} 
OT can be seen as the search for a plan that moves (transports) a probability measure $\mu_1$ onto another measure $\mu_2$ with a minimum cost measured by some function $c$. 
In our case, we use the squared Euclidean distance $L^2_2$, but other domain-specific measures, more suited to the problem at hand, could be used instead. 
In the relaxed formulation of Kantorovitch~\citep{Kantorovich42}, OT seeks for an optimal coupling that can be seen as a joint probability distribution between $\mu_1$ and $\mu_2$. In other words,
if we define $\Pi(\mu_1,\mu_2)$ as the space of  probability distributions over $\mathbb{R}^2$ with marginals $\mu_1$ and $\mu_2$, the optimal transport is the coupling $\ga \in \Pi(\mu_1,\mu_2)$,
which minimizes the following quantity:
  \begin{equation*}
 W_c(\mu_1,\mu_2) = \inf_{\ga \in \Pi(\mu_1,\mu_2)} \int_{\mathbb{R}^2} c(\x_1,\x_2)d\ga(\x_1,\x_2),
 \end{equation*}
 where $c(\x_1,\x_2)$ is the cost of moving $\x_1$ to $\x_2$ (drawn from distributions $\mu_1$ and $\mu_2$, respectively). 
In the discrete versions of the problem, {\em i.e.} when $\mu_1$ and $\mu_2$ are defined as empirical measures based on vectors in $\mathbb{R}^d$, $\Pi(\mu_1,\mu_2)$ denotes the polytope of matrices $\ga$ such that $\ga\bm{1} = \mu_1, \ga^T\bm{1} = \mu_2$ and the previous equation reads:
 \begin{equation}
 W_\C(\mu_1,\mu_2) = \min_{\gamma \in \Pi(\mu_1,\mu_2)}  \langle \ga, \C \rangle_F,
 \label{eq:kanto}
 \end{equation}
 where $\langle\cdot, \cdot\rangle_F$ is the Frobenius dot product, $\C \geq 0$ is a cost matrix $\in \mathbb{R}^{n_1\times n_2}$, representing the pairwise costs of transporting bin $i$ to bin $j$, and $\ga$ is a joint distribution given by a matrix of size  $n_1\times n_2$,
 with marginals defined as $\mu_1$ and $\mu_2$. 
Solving equation~\eqref{eq:kanto} is a simple linear programming problem with equality constraints, but its dimensions scale quadratically with the size of the sample. Alternatively, one can consider a regularized version of the problem, which has the extra benefit of being faster to solve.    

\subsection{Entropic Regularization}
\label{subsec:opreg}
In \citep{cuturi:2013}, the authors added a regularization term to $\ga$ that controls the smoothness of the coupling through the entropy of $\ga$. 
The entropy regularized version of the discrete OT reads:
\begin{equation}
 W_{\C,\epsilon}(\mu_1,\mu_2) = \min_{\ga \in \Pi(\mu_1,\mu_2)}  \langle\ga, \C\rangle_F - \epsilon h(\ga),
\label{eq:regTO}
\end{equation}
where $h(\ga)= - \sum_{ij} \ga_{ij} (\log \ga_{ij}-1)$ is the entropy of $\ga$. 
Similarly, denoting the Kullback-Leibler divergence ($\kldiv$) as
$\kldiv(\ga|\rho) =  \sum_{ij} \ga_{ij} (\log \frac{\ga_{ij}}{\rho_{ij}}-1) =  \langle\ga, \log \frac{\ga}{\rho} - \one\rangle_F$, one can establish the following link between OT and Bregman projections.

\begin{proposition} {\bf 
~\citep[Eq. (6,7)]{benamou:2015}.}
For $\zeta=\exp\left(-\frac{C}{\epsilon}\right)$, the minimizer  $\ga^\star$  of \eqref{eq:regTO}  is the solution of the following Bregman projection
\begin{equation*}
\ga^\star = \argmin_{\ga \in \Pi(\mu_1,\mu_2)} \kldiv(\ga|\zeta).
\label{eq:breg}
\end{equation*}
For an undefined $\mu_2$, $\ga^\star$ is solely constrained by the marginal $\mu_1$ and is the solution of the following closed-form projection:
\begin{equation}
\ga^\star = \diag\left(\frac{\mu_1}{\zeta\one}\right)\zeta,
\label{eq:closeBreg}
\end{equation}
where the division has to be understood component-wise. 
\end{proposition}
As it follows from this proposition, the entropy regularized version of OT can be solved with a simple algorithm based on successive projections over the two marginal constraints and admits a closed form solution.
We refer the reader to~\citep{benamou:2015} for more details on this subject. 

\subsection{Application to Domain Adaptation}
A solution to the two domains adaptation problem based on OT has been proposed in \citep{courty14}. It consists in estimating a transformation of the source domain sample that minimizes their average displacement w.r.t. target sample, \emph{i.e.} an optimal transport solution between the discrete distributions of the two domains. 
The success of the proposed algorithm is due to  an important advantage offered by OT metric over other distances used in DA (e.g. MMD): it preserves the topology of the data and admits a rather efficient estimation. The authors further added a regularization term 
used to encourage instances from the target sample to be transported to instances of the source sample of the same class, therefore promoting group sparsity in $\gamma$ thanks to the $\Vert \cdot \Vert^p_q$ norm with $q = 1$ and $p = \frac{1}{2}$ \citep{courty14} or $q = 2$ and $p = 1$ \citep{DBLP:journals/pami/CourtyFTR17}. 

\section{DOMAIN ADAPTATION UNDER THE TARGET SHIFT}
\label{sec:limits}
In this section, we formalize the target shift problem and provide a generalization bound that shows the key factors that have an impact when learning under it. 

To this end, let us consider a binary classification problem with $K$ source domains, each being represented by a sample of size $n^{(k)}, k = 1, \dots, K$, drawn from a probability distribution $P^{k}_S = (1-\pi^k_S)P_0 + \pi^k_SP_1$. Here $0 < \pi^k_S < 1$ and $P_0, P_1$ are marginal distributions of the source data given the class labels $0$ and $1$, respectively with $P_0 \neq P_1$. 
We also possess a target sample of size $n$ drawn from a probability distribution $P_T= (1-\pi_T)P_0 + \pi_T P_1$ such that  $\exists \ j \in [1, \dots, N]: \pi_S^j \neq \pi_T$. This last condition is a characterization of target shift used in previous theoretical works on the subject \citep{pmlr-v30-Scott13}. 

Following \citep{bendavid10}, we define a domain as a pair consisting of a distribution $P_D$ on some space of inputs $\Omega$ 
and a labeling function $f_D: \Omega \rightarrow [0,1]$. A hypothesis class $\mathcal{H}$ is a set of functions so that $\forall h \in \mathcal{H}, h: \Omega \rightarrow \lbrace0,1\rbrace$. Given a convex loss-function $l$, the true risk with respect to the distribution $P_D$, for a labeling function $f_D$ (which can also be a hypothesis) and a hypothesis $h$ is defined as
\begin{equation}
\epsilon_D (h,f_D) = \mathbb{E}_{x \sim P_D} \left[l(h(x),f_D(x))\right].
\end{equation}

In the multi-source case, when the source and target error functions are defined w.r.t. $h$ and $f_S^{(k)}$ or $f_T$, we use the shorthand $\epsilon_S^{(k)} (h, f_S^{(k)}) = \epsilon_S^{(k)} (h)$ and $\epsilon_T (h, f_T) = \epsilon_T (h)$. The ultimate goal of multi-source DA then is to learn a hypothesis $h$ on $K$ source domains that has the best possible performance in the target one.

To this end, we define the combined error of source domains as a weighted sum of source domains error functions:
\begin{equation*}
\epsilon_S^{\bm{\alpha}} = \sum_{k=1}^K \alpha_k\epsilon_S^{(k)}, \sum_{k=1}^K \alpha_k = 1, \alpha_k\in [0,1] \ \forall k \in [1, \dots, K].
\end{equation*}
We further denote by $f_S^{\bm{\alpha}}$ the labeling function associated to the distribution mixture $P_S^{\bm{\alpha}} = \sum_{k=1}^N \alpha_j P_S^j$. In multi-source scenario, the combined error is minimized in order to produce a hypothesis that is used on the target domain. Here different weights $\alpha_k$ can be seen as measures reflecting the proximity of the corresponding source domain distribution to the target one.

For the target shift setup introduced above, we can prove the following proposition.
\begin{proposition}\label{th:bound}
Let $\mathcal{H}$ denote the hypothesis space of predictors $h: \Omega \rightarrow \lbrace0,1\rbrace$ and $l$ be a convex loss function. Let $\text{disc}_l(P_S,P_T) = \max_{h,h' \in \mathcal{H}} \vert \epsilon_S(l(h,h')) - \epsilon_T(l(h,h'))\vert$ be the discrepancy distance~\citep{DBLP:conf/colt/MansourMR09} between two probability distributions $P_S$ and $P_T$. 
Then, for any fixed $\bm{\alpha}$ and for any $h \in \mathcal{H}$ the following holds:
\begin{align*}
\epsilon_T(h) \leq \epsilon_S^{\bm{\alpha}}(h) + \vert \pi_T - \sum_{j=1}^N \alpha_j \pi_S^j \vert \text{disc}_l(P_0, P_1)+\lambda,
\end{align*}
where
$\lambda = \underset{h \in \mathcal{H}}{\min} \ \epsilon_{S}^{\bm{\alpha}}(h)+\epsilon_T(h)$ represents the joint error between the combined source error and the target one\footnote{Proofs of several theoretical results of this paper can be found in the Supplementary material.}. 
\end{proposition}
The second term in the bound can be minimized for any $\alpha_k$ when $\pi_T=\pi_S^k,\ \forall k$. This can be achieved by using a proper reweighting of the class distributions in the source domains, but requires to have access to the target proportion which is assumed to be unknown. In the next section, we propose to estimate optimal proportions by minimizing the sum of the Wasserstein distances between all reweighted sources and the target distribution. In order to justify this idea, we prove below that the minimization of the Wasserstein distance between a weighted source distribution and a target distribution yields the optimal proportion estimation. To proceed, let us consider the multi-class problem with $C$ classes, where the target distribution is defined as 
$$P_T=\sum_{i=1}^C \pi_i^T P_i,$$
with $P_i$ being a distribution of class $i \in \{1, \dots, C\}$. As before, the source distribution with weighted classes can be then defined as 
$$P_S^\pi=\sum_i \pi_i P_i,$$ 
where $\pi \in \Delta_C$ are coefficients lying in the probability simplex $\Delta_C \eqdefU \{\alpha \in \mathbb{R}^C_+: \sum_{i=1}^C \alpha_i = 1\}$ that reweigh the corresponding classes.

As the proportions of classes in the target distribution are unknown, our goal is to reweigh source classes distributions by solving the following optimization problem:
\begin{align}
	\pi^\star=\argmin_{\pi \in \Delta_C}\ W(P^\pi_S,P_T).
    \label{optim_prob}
\end{align}
We can now state the following proposition.
\begin{proposition}
Assume that $\forall i, \nexists \alpha \in \{\Delta_C|\alpha_i=0, P_i=\sum_j \alpha_j P_j\}$. Then, for any distribution $P_T$, the unique solution $\pi^*$ minimizing (\ref{optim_prob}) is given by $\pi^T$.
\end{proposition}
Note that this result extends straightforwardly to the multi-source case where the optimal solution of minimizing the sum of the Wasserstein distance for all source distributions is the target domain proportions. As real distributions are accessible only through available finite samples, in practice, we propose to minimize the Wasserstein distance between the empirical target distribution $\hat P_T$ and the empirical source distributions $\hat P_S^{k}$. The convergence of the exact solution of this problem with empirical measures can be characterized using the concentration inequalities established for Wasserstein distance in \citep{bobkov,fournier:hal-00915365} where the rate of convergence is inversely proportional to the number of available instances in source domains and consequently, to the number of source domains.


\section{JOINT CLASS PROPORTION AND OPTIMAL TRANSPORT (JCPOT)}
\label{sec:JCPOT}
In this section, we introduce the proposed JCPOT method, that aims at finding the optimal transportation plan and estimating class proportions jointly. The main underlying idea behind JCPOT is to reweigh instances in the source domains in order to compensate for the discrepancy between the source and target domains class proportions.


\subsection{Data and Class-Based Weighting}
We assume to have access to several data sets corresponding to $K$ different domains $\Xsk$, $k=1,\hdots,K$. These domains are formed by $\nk$ instances $\xsi \in \mathbb{R}^d$ with each instance being associated with one of the $C$ classes of interest. {In the following, we use the superscript $(k)$ when referring to quantities in one of the source domains (e.g $\mu^{(k)}$) and the equivalent without superscript when referring to the same quantity in the (single) target domain (e.g. $\mu$).}
Let $\ysi$ be the corresponding class, {\em i.e.} $\ysi \in \{1,\hdots,C\}$.  
We are also given a target domain $\X$, populated by $\nt$ instances defined in $\mathbb{R}^d$. The goal of unsupervised multi-source adaptation is to recover the classes $\yi$ of the target domain samples, which are all unknown. 

JCPOT works under the target shift assumption presented in Section~\ref{sec:limits}. For every source domain, we assume that its data points follow a probability distribution function or probability measure $\muk$ ($\int \muk =1$). 
In real-world situations, $\muk$ is only accessible through the instances $\xsi$ that we can use to define a distribution $\muk = \sum^\nk_{i=1} \mik \dik$, where $\dik$ are Dirac measures located
at $\xsi$, and $\mik$ is an associated probability mass. By denoting the corresponding vector of mass as $\mk$, {\em i.e.} $\mk = [\mik]_{i=\{1,\hdots,\nk\}}$, and 
$\dXk$ the corresponding vector of Dirac measures, one can write $\muk = (\mk)^T \dXk$. Note that when the data set is a collection of independent data points, the weights of all instances in the sample are usually set to be equal. In this work, however, we use different weights for each class of the source domain so that we can adapt the proportions of classes w.r.t. the target domain.
To this end, we note that the measures can be decomposed among the $C$ classes as $\muk = \sum^C_{c=1} \muck$. We denote by $\hck = \int \muck$ the proportion of class $c$ in $\Xsk$. By construction, we have $\hck = \sum^\nk_{i=1} \delta(\ysi = c) \mik$. 

Since we chose to have equal weights in the classes, we define two linear operators $\Dmk \in \mathbb{R}^{C\times \nk}$ and $\Dhk \in \mathbb{R}^{\nk \times C}$ that allow to express the transformation from the vector of mass $\mk$ to the class proportions $\hk$ and back:
\[\Dmk(c,i) = \left\{ \begin{array}{l l}1 & \quad \text{if $\ysi= c$,}\\0 & \quad \text{otherwise,}\\ \end{array} \right.\]
 and \[\Dhk(i,c) = \left\{ \begin{array}{l l}\frac{1}{\#\{\ysi=c\}_{i=\{1,\hdots,\nk\}}} & \quad \text{if $\ysi=c$,}\\0 & \quad \text{otherwise.}\\ \end{array} \right. \]
$\Dmk$ allows to retrieve the class proportions with $\hk = \Dmk \mk$ and $\Dhk$ returns weights for all instances for a given vector of class proportions with $\mk = \Dhk \hk$, where the masses are distributed equiproportionnally among all the data points associated to one class. For example, for a source domain with 5 elements from which first 3 belong to class 1 and the other to class 2, 
\[\mathbf{D}_1 = \begin{bmatrix}
    1 & 1 & 1 & 0 & 0 \\
    0 & 0 & 0 & 1 & 1
\end{bmatrix},\] 
$\mathbf{m} = [\frac{1}{5},\frac{1}{5},\frac{1}{5},\frac{1}{5},\frac{1}{5}]^T$ so that $\mathbf{h} = \mathbf{D}_1\mathbf{m} = [\frac{3}{5},\frac{2}{5}]$. These are the class proportions of this source domain. On the other hand, 
\[\mathbf{D}_2 = \begin{bmatrix}
    \frac{1}{3} & \frac{1}{3} & \frac{1}{3} & 0 & 0 \\
    0 & 0 & 0 & \frac{1}{2} & \frac{1}{2}
\end{bmatrix}^T,\] 
so that $\mathbf{D}_2\mathbf{h} = \mathbf{m}$ and $\mathbf{D}_1\mathbf{D}_2 = \bm{I}$.

\subsection{Multi-Source Domain Adaptation with {JCPOT}}

As illustrated in Section~\ref{sec:intro}, having matching proportions between the source and the target domains helps in finding better couplings, and, as shown in Section~\ref{sec:limits} it also enhances the adaptation results. 

To this end, we propose to estimate the class proportions in the target domain by solving a constrained Wasserstein barycenter problem~\citep{benamou:2015} for which we use the operators defined above to match the proportions to the uniformly weighted target distribution. The corresponding optimization problem can be written as follows: 
\begin{equation}
\argmin_{\htar \in \Delta_C}\quad \sum_{k=1}^K \lambda_k W_ {\epsilon,C^{(k)}}\left((\Dhk \htar)^T\dXk, \mu\right),
\label{eq:Wbar}
\end{equation}
where regularized Wasserstein distances are defined as
\begin{equation*}
W_ {\epsilon,C^{(k)}}(\mu^{(k)}, \mu) \eqdefU
\min_{\ga^{(k)} \in \Pi(\mu^{(k)}, \mu)} \kldiv(\ga^{(k)}|\zeta^{(k)}),
\label{eq:defW}
\end{equation*}
provided that $\zeta^{(k)}=\exp\left(-\frac{C^{(k)}}{\epsilon}\right)$ with $\lambda_k$ being convex coefficients $\left(\sum_k \lambda_k =1\right)$ accounting for the relative importance of each domain. Here, we define the set $\Gamma=\{\ga^{(k)}\}_{k=1\hdots K} \in (\mathbb{R}^{\nk\times \nt})^K$ {as the set of couplings between each source and the target domains}. This problems leads to $K$ marginal constraints $ \ga^T_k \one_\nt=\one_n / n$ w.r.t. the uniform target distribution, and $K$ marginal constraints $ \Dmk \ga_k \one_\nt= \htar$ related to the unknown proportions $\htar$. 

Optimizing for the first $K$ marginal constraints can be done independently for each $k$ by solving the problem expressed in Equation~\ref{eq:closeBreg}. On the contrary, the remaining $K$ constraints require to solve the proposed optimization problem for $\Gamma$ and $\htar$, simultaneously. To do so, we formulate the problem as a Bregman projection with prescribed row sum ($\forall k \;\; \Dmk \ga^{(k)} \one_\nt=\ \htar$), \ie, 
\begin{equation}
\begin{aligned}
\htar^\star &= \argmin_{\htar \in \Delta_C,\Gamma} \sum_{k=1}^K \lambda_k \kldiv(\ga^{(k)}|\zeta^{(k)}) \\
&\textrm{s.t.} \;\;\; \forall k \;\; \Dmk \ga^{(k)} \one_\nt= \htar. 
\label{eq:projsol}
\end{aligned}
\end{equation} 
This problem admits a closed form solution that we establish in the following result.
\begin{proposition} 
The solution of the projection defined in Equation~\ref{eq:projsol} is given by:
\begin{align}
\forall k,  \ga_k = \diag\left(\frac{\Dhk \htar}{\zeta^{(k)}\one_\nt}\right)\zeta^{(k)}, \htar=\Pi_{k=1}^K (\Dmk (\zeta^{(k)} \one_\nt) ) ^{\lambda_k}.\nonumber
\label{eq:closeBreg2}
\end{align}
\end{proposition}
The initial problem can now be solved through an Iterative Bregman projections scheme summarized in Algorithm~\ref{alg1}. Note that the updates for coupling matrix 
in lines 5 and 7 of the algorithm can be computed in parallel for each domain.
\begin{figure}[!t]
\vspace{-2mm}
\centering
   \begin{minipage}[h]{\linewidth}
        \begin{algorithm}[H]
        \caption{{Joint Class Proportion and Optimal Transport (JCPOT)}}
        \label{alg1}
        \begin{algorithmic}[1]
        \STATE \textbf{Input:} $\epsilon$, maxIter, $\forall k$ ($\C^{(k)}$ and $\lambda^{(k)}$)
        \STATE $cpt \leftarrow 0$,   
        \STATE $err \leftarrow \infty$
        \FORALL{$k=1,\hdots, K$}
        \STATE $\zeta^{(k)} \leftarrow \exp(-\frac{\C^{(k)}}{\epsilon})$, $\quad \forall \, k$
        \ENDFOR
        \WHILE {$cpt < $ maxIter {\bf and} $err > $ threshold } 
        	\FORALL{$k=1,\hdots, K$}
        	 \STATE $\zeta^{(k)} \leftarrow \diag(\frac{\mt}{\zeta^{(k)}\one})\zeta^{(k)}$, $\quad \forall \, k$
        	\ENDFOR
                \STATE $\htar^{(cpt)}  \leftarrow \exp(\sum^K_{k=1}\lambda^{(k)} \log(\Dmk \zeta^{(k)}\one))$
             \FORALL{$k=1,\hdots, K$}
        	\STATE $\zeta^{(k)} \leftarrow \zeta^{(k)} \diag(\frac{\Dhk\htar}{\zeta^{(k)}\one})$, $\quad \forall \, k$
        	\ENDFOR
        	\STATE err $\leftarrow || \htar^{(cpt)} - \htar^{(cpt-1)}||_2$,  
         \STATE   cpt $\leftarrow$ cpt + 1	
        \ENDWHILE 
        \RETURN \htar, $\forall k$ $\zeta^{(k)}$ 
        \end{algorithmic}
        \end{algorithm}
    \end{minipage}
\end{figure}
\subsection{Classification in the Target Domain}
\label{subsec:classif}
When {both the class proportions and} the corresponding coupling matrices are obtained, we need to adapt source and target samples and classify unlabeled target instances. Below, we provide two possible ways that can be used to perform these tasks.

\paragraph{Barycentric mapping} In \citep{DBLP:journals/pami/CourtyFTR17}, the authors proposed to use the OT matrices to estimate {the position of each source instance} as the barycenter of {the} target instances, weighted by the mass from the source sample.
 This approach extends to multi-source setting and {naturally provides} a target-aligned 
position for each point {from each source domain}. These adapted source samples can then be used to learn a classifier and apply it directly on the target sample. In the sequel, we denote the variations of JCPOT that use the barycenter mapping as \texttt{JCPOT-PT}. For this approach, \citep{DBLP:journals/pami/CourtyFTR17} noted that too much regularization has a shrinkage effect on {the new positions, since} the mass spreads to all target points in this configuration. Also, it requires the estimation of a target classifier{, trained on the transported source samples,} to provide {predictions for} the target sample. 
\paragraph{Label propagation} We propose alternatively to use the OT matrices to perform label propagation onto the target sample. Since we have access to the labels in the source domains and {since} the OT matrices {provide} the transportation of mass, we can measure for each target instance the proportion of mass coming from every class. Therefore, we propose to estimate the label proportions {for the} target sample with $\mathbf{L}=\sum_{k=1}^K\lambda_k\Dmk\ga^{(k)}$
where the component $l_{c,i}$ {in \textbf{L}} contains the probability estimate of target sample $i$ to {belong} to class $c$. Note that this label propagation technique can be seen as boosting, since the expression of $\mathbf{L}$ {corresponds to} a linear combination of weak classifiers from each source domain. To the best of our knowledge, this is the first time such type of approach is proposed in DA. In the following, we denote it by \texttt{JCPOT-LP} where LP stands for label propagation.

\section{EXPERIMENTAL RESULTS}
\label{sec:experiments}
In this section, we present the results of our algorithm for both synthetic and real-world data from the task of remote sensing classification.

\paragraph{Baseline and state-of-the-art methods} 
We compare the proposed JCPOT algorithm to three other methods designed to tackle target shift, namely \texttt{betaEM}, the variation of the EM algorithm proposed in \citep{Chan:2005:WSD} 
\texttt{betaKMM}, an algorithm based on the kernel embeddings proposed in \citep{conf/icml/ZhangSMW13}
\footnote{code available at \url{http://people.tuebingen.mpg.de/kzhang/Code-TarS.zip}.}
{, and \texttt{MDA Causal}, a multi-domain adaptation strategy with a causal view~\citep{zhang2015multi} \footnote{code available at \url{https://mgong2.github.io/papers/MDAC.zip}}}. \rf{Note that despite the existence of several deep learning methods that deal with covariate shift, e.g.~\citep{GaninL15}, to the best of our knowledge, none of them tackle specifically the problem of target shift.
}

As explained in \ref{subsec:classif}, our algorithms can obtain the target labels in two different ways, either based on label propagation (\texttt{JCPOT-LP}) or based on transporting points and applying a standard classification algorithm after transformation (\texttt{JCPOT-PT}). Furthermore, we also consider two additional DA algorithms that use OT \citep{courty14}: \texttt{OTDA-LP} and \texttt{OTDA-PT} that align the domains based on OT but without considering the discrepancies in class proportions. 

\subsection{Synthetic Data}
\label{sec:synthetic}

\paragraph{Data generation}

In the multi-source setup, we sample 20 source domains, each consisting of 500 instances and a target domain with 400 instances. We vary the source domains' class proportions randomly while keeping the target ones equal to $[0.2;0.8]$. For more details on the generative process and some additional empirical results regarding the sensitivity of \texttt{JCPOT} to hyper-parameters tuning and the running times comparison, we refer the reader to the Supplementary material.

\paragraph{Results}  Table~\ref{tab:multi_source_sim} gives average performances over five runs for each domain adaptation task when the number of source domains varies from $2$ to $20$. As \texttt{betaEM}, \texttt{betaKMM} and \texttt{OTDA} are not designed to work in the multi-source scenario, we fusion the data from all source domains and use it as a single source domain. From the results, we can see that the algorithm with label propagation (\texttt{JCPOT-LP}) provides the best results and outperforms other state-of-the-art DA methods, {except for 20 source domains, where \texttt{MDA Causal} slightly surpasses our method. It is worth noting that}, all methods addressing specifically the target shift problem perform better that the \texttt{OTDA} method designed for covariate shift. This result justifies our claim about the necessity of specially designed algorithms that take into account the shifting class proportions in DA. 

On the other hand, we also evaluate the accuracy of proportion estimation of our algorithm and compare it with the results obtained by the algorithm proposed in~\cite{pmlr-v30-Scott13}\footnote{code available at \url{http://web.eecs.umich.edu/~cscott/code/mpe_v2.zip}}. As this latter was designed to deal with binary classification, we restrict ourselves to the comparison on simulated data only and present the deviation of the estimated proportions from their true value in terms of the $L^1$ distance in Table \ref{tab:prop_acc}.  
\renewcommand*{\arraystretch}{1.2}
\begin{table}[!t]
\centering
\resizebox{\linewidth}{!}{
\begin{tabular}{c|ccccccc}
\hline
&\multicolumn{7}{c}{Number of source domains}\\
& 2 & 5 & 8 & 11 & 14 & 17 & 20\\
\hline
JCPOT & 0.039 & 0.045 & \textbf{0.027} & \textbf{0.029} & \textbf{0.035}  & \textbf{0.033} & \textbf{0.034}  \\
\cite{pmlr-v30-Scott13} & \textbf{0.01}  &  \textbf{0.044}  &  0.06  &  0.10  &  0.22 & \textbf{0.033} & 0.14\\
\hline
\end{tabular}
}
\caption{Accuracy of proportions estimation for simulated data.}
\label{tab:prop_acc}
\end{table}
From this Table, we can see that our method gives comparable or better results in most of the cases. Furthermore, our algorithm provides coupling matrices that allow to align the source and target domains samples and to directly classify target instanes using the label propogation described above.

\setlength{\tabcolsep}{2pt}
\renewcommand*{\arraystretch}{1.15}
\begin{table*}[!t]
\centering
\resizebox{0.95\textwidth}{!}{
\begin{tabular}{|ccc|cccccccc|c|}
\hline
\parbox[c][1cm][c]{1.7cm}{\small  \centering \# of source\\ domains} & \parbox[c][1cm][c]{2cm}{\small \centering Average class \\ proportions} & \parbox[c][1cm][c]{1.5cm}{\small \centering \# of source \\instances} &  \parbox[c][1cm][c]{1.5cm}{\small \centering No adaptation} &  \parbox[c][1cm][c]{1.5cm}{\centering \texttt{OTDA} \\ \texttt{PT}} & \parbox[c][1cm][c]{1.5cm}{\centering \texttt{OTDA} \\ \texttt{LP}} & \parbox[c][1cm][c]{1.5cm}{\centering \texttt{beta} \\ \texttt{EM}} &   \parbox[c][1cm][c]{1.5cm}{\centering \texttt{beta} \\ \texttt{KMM}} & 
\parbox[c][1cm][c]{1.5cm}{\centering \texttt{MDA} \\ \texttt{Causal}} & 
\parbox[c][1cm][c]{1.5cm}{\centering \texttt{JCPOT} \\ \texttt{PT}} &  \parbox[c][1cm][c]{1.5cm}{\centering \texttt{JCPOT} \\ \texttt{LP}}  &   \parbox[c][1cm][c]{1.5cm}{\centering \small  Target \\ only}\\
\hline
\hline
\multicolumn{9}{c}{Multi-source simulated data}\\
\hline
\hline
2 & $[ 0.64 \ 0.36]$ & 1000 & 0.839 & 0.75 & 0.69 & 0.82 & \underline{0.86} & \underline{0.86} & 0.78 & {\bf 0.87} & 0.854 \\
5 & $[ 0.5 \ 0.5]$ & 2'500 & 0.80 & 0.63 & 0.74 & 0.84 & 0.85 & \underline{0.866}  & 0.813 &  \textbf{0.878} & 0.854\\
8 & $[ 0.47 \ 0.53]$ & 4'000 & 0.79 & 0.75 & 0.65 & 0.85 & 0.85 & \underline{0.866} & 0.78 & \textbf{0.88} & 0.854 \\
11 & $[ 0.48 \ 0.52]$ &  5'500 & 0.81 & 0.53 & 0.76 & 0.83 &  0.85 & \underline{0.867} & 0.8 & \textbf{0.874} & 0.854\\
14 & $[ 0.53 \ 0.47]$ & 7'000 & 0.83 & 0.70 & 0.75 & \underline{0.87} & 0.86 & 0.85 & 0.77 &  \textbf{0.88} & 0.854 \\
17 & $[ 0.52  \ 0.48]$ & 8'500  & 0.82 & 0.75 & 0.76  & \underline{0.86} & \underline{0.86} & \underline{0.86} & 0.79 &  \textbf{0.878} & 0.854 \\
20 & $[ 0.51  \ 0.49]$ & 10'000  & 0.80 & 0.77 & 0.79 & 0.87 & 0.854 & \textbf{0.877} & 0.86 & \underline{0.874} & 0.854 \\
\hline
\hline
\multicolumn{9}{c}{Zurich data set}\\
\hline
\hline
2 & $[ 0.168 \ 0.397 \ 0.161 \ 0.273]$ & 2'936 & 0.61 & 0.52 &  0.57 & 0.59 & 0.61 & \underline{0.65} &0.59 & {\bf 0.66} & 0.65 \\
5 & $[ 0.222 \ 0.385  \ 0.181 \ 0.212]$ & 6'716 & 0.62 & 0.55 &  0.6 & 0.58 & 0.6  & \underline{0.66} &0.58 &  \textbf{0.68} & 0.64\\
8 & $[ 0.248 \ 0.462 \ 0.172 \ 0.118]$ & 16'448 & 0.63 & 0.54 &  0.59 & 0.59 & 0.61 & \underline{0.67} & 0.63 & \textbf{0.71} & 0.65 \\
11 & $[ 0.261 \ 0.478 \ 0.164 \ 0.097]$ &  21'223 & 0.63 & 0.54 &  0.58 & 0.59 &  0.62 & \underline{0.67} &0.58 & \textbf{0.72} &0.673\\
14 & $[ 0.256 \ 0.448 \ 0.192 \ 0.103]$ & 27'875 & 0.63 & 0.52 &  0.58 & 0.59 & 0.62 & \underline{0.67} &0.59 &  \textbf{0.72} & 0.65 \\
17 & $[ 0.25  \ 0.415 \ 0.207 \ 0.129]$ & 32'660  & 0.63 & 0.5 &  0.59 & 0.59 & 0.63 & \underline{0.67} &0.6 &  \textbf{0.73} & 0.61 \\
\hline
\end{tabular}}
\caption{Results on multi-source simulated data and pixel classification results obtained on the Zurich Summer data set. The underline numbers and those in bold correspond to the second and the best performances obtained for each configuration, respectively.
}\label{tab:multi_source_sim}
\end{table*}
\subsection{Real-World Data From Remote Sensing Application}
\label{sec:real}
\begin{figure}
\centering
\begin{minipage}{0.3\linewidth}
  \includegraphics[width=\linewidth]{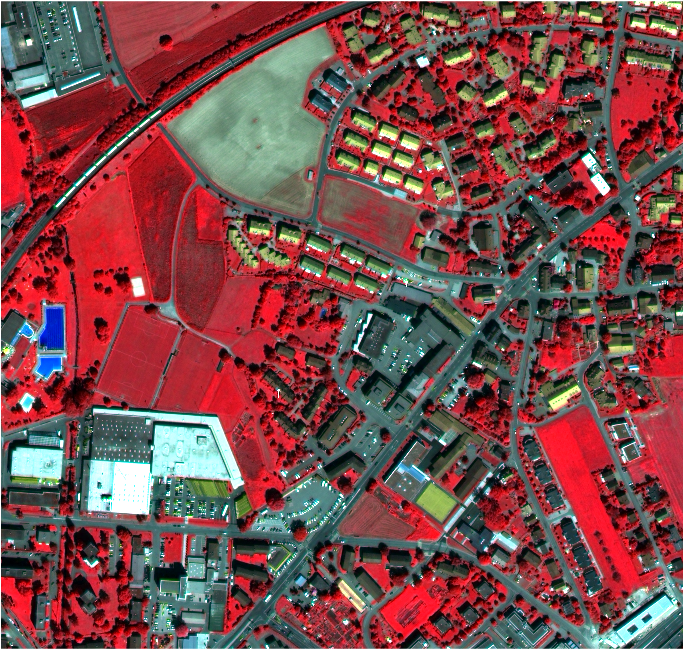}
  \end{minipage} 
\begin{minipage}{0.3\linewidth}
  \includegraphics[width = \linewidth]{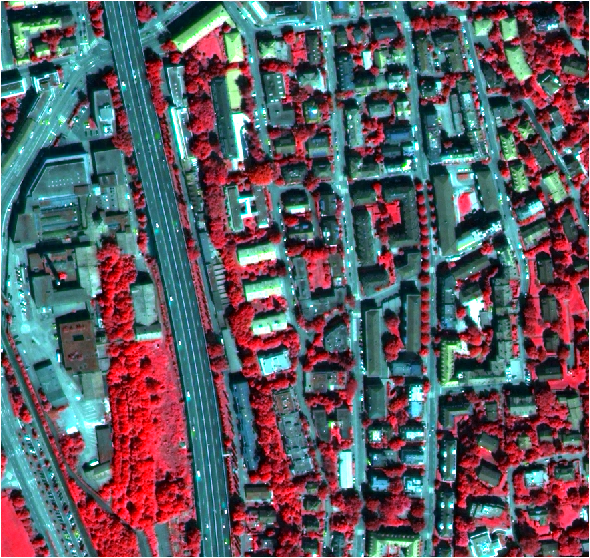}
\end{minipage}
\begin{minipage}{0.3\linewidth}
  \includegraphics[width = \linewidth]{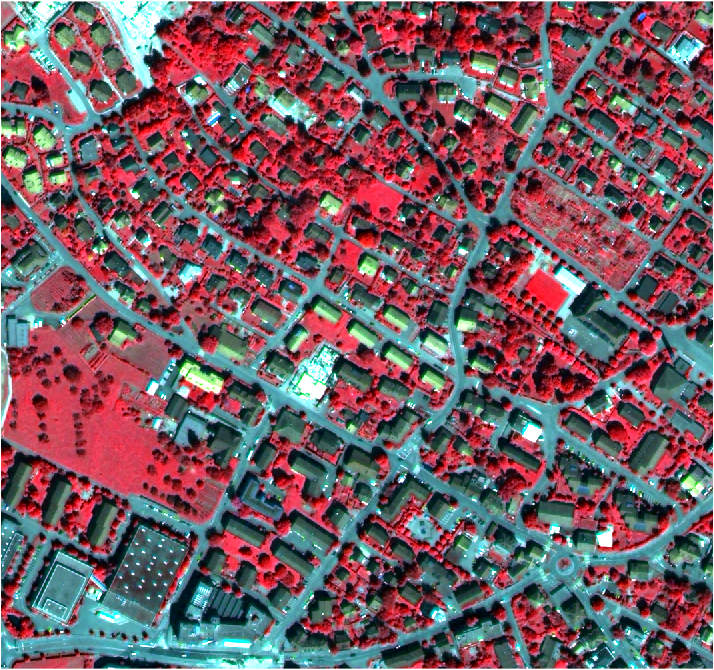}
\end{minipage}
\\
\begin{minipage}{0.3\linewidth}
  \includegraphics[width=\linewidth]{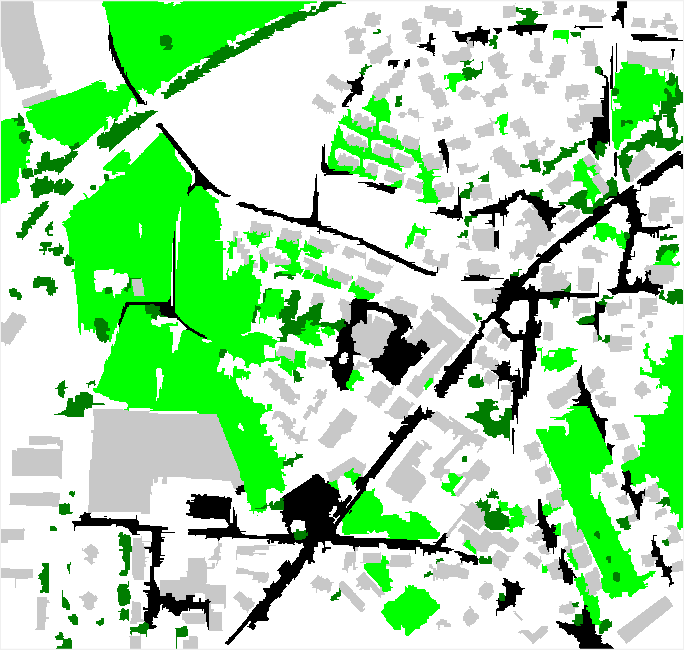}
  \end{minipage} 
\begin{minipage}{0.3\linewidth}
  \includegraphics[width = \linewidth]{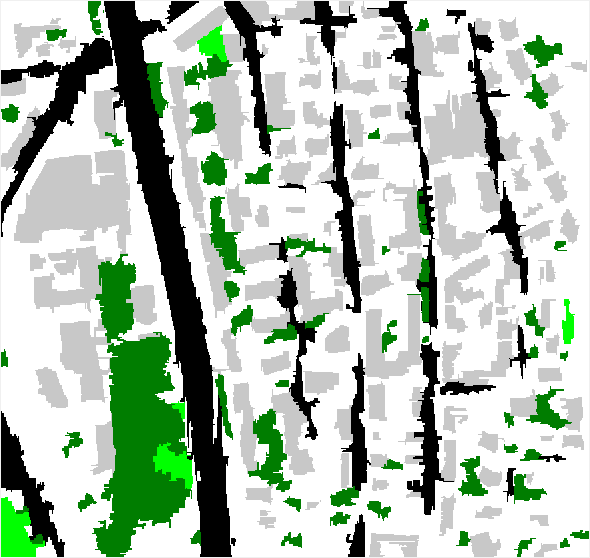}
\end{minipage}
\begin{minipage}{0.3\linewidth}
  \includegraphics[width = \linewidth]{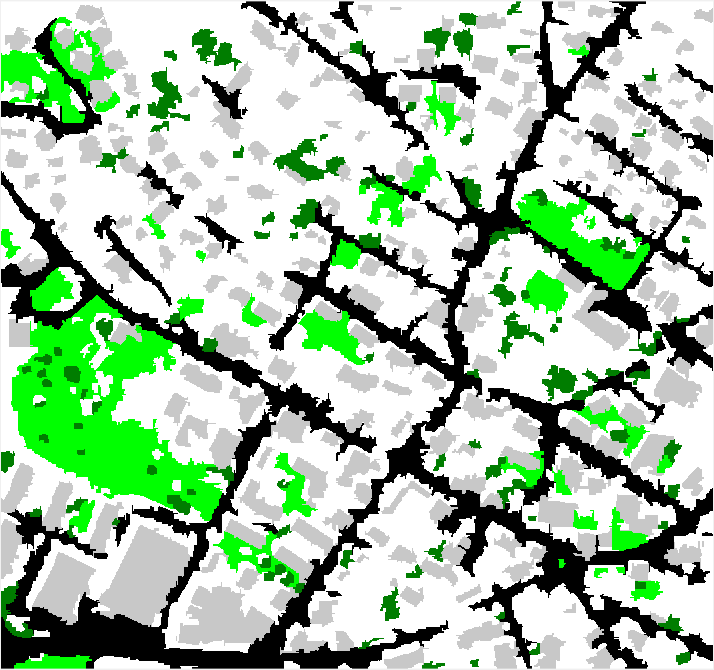}
\end{minipage}  
  \caption{Original (top row) and ground truths (bottom row) images from Zurich data set. Class proportions are highly imbalanced between the images. Color legend for ground truths: black: roads, gray: buildings, green: grass; dark green: trees.}
  \label{fig:multi_source_zurich}
\end{figure}

\paragraph{Data set} We consider the task of classifying superpixels from satellite images at very high resolution into a set of land cover/land use classes~\citep{tuia15}.
We use the `Zurich Summer' data set\footnote{https://sites.google.com/site/michelevolpiresearch/data\-/zurich-dataset}, composed of 20 images issued from a large image acquired by the QuickBird satellite over the city of Zurich, Switzerland in August 2002 where the features are extracted as described in \citep[Section 3.B]{tuia_zurich}. For this data set, we consider a multi-class classification task corresponding to the classes Roads, Buildings, Trees and Grass shared by all images. The number of superpixels per class is imbalanced and varies across images: thus it represents a real target shift problem. We consider 18 out of the 20 images, since two images exhibit a very scarce ground truth, making a reliable estimation of the true classes proportions difficult. We use each image as the target domain (average class proportions with standard deviation are $[0.25\pm 0.07, \ 0.4\pm 0.13, \ 0.22\pm 0.11, \ 0.13\pm 0.11]$) while considering remaining 17 images as source domains. Figure \ref{fig:multi_source_zurich} presents both the original and the ground truths of several images from the considered data set. One can observe that classes of all three images have very unequal proportions compared to each other.

\paragraph{Results}
The results over 5 trials obtained on this data set are reported in the lower part of Table \ref{tab:multi_source_sim}. The proposed \texttt{JCPOT} method based on label propagation significantly improves the classification accuracy over the other baselines. The results show an important improvement over the ``No adaptation'' case, with an increase reaching 10\% for \texttt{JCPOT-LP}. We also note that the results obtained by \texttt{JCPOT-LP} outperform the ``Target only" baseline. This shows the benefit brought by multiple source domains as once properly adapted, they represent a much larger annotated sample that the target domain sample alone. This claim is also confirmed by an increasing performance of our approach with the increasing number of source domains. Overall, the obtained results show that the proposed method handles the adaptation problem quite well and thus allows to avoid manual labeling in real-world applications.

\section{CONCLUSIONS}
In this paper we proposed JCPOT, a novel method dealing with target shift: a particular and largely understudied DA scenario occurring when the difference in source and target distributions is induced by differences in their class proportions. To justify the necessity of accounting for target shift explicitly, we presented a theoretical result showing that unmatched proportions between source and target domains lead to inefficient adaptation. Our proposed method addresses the target shift problem by tackling the estimation of class proportions and the alignment of domain distributions jointly in optimal transportation framework. We used the idea of Wasserstein barycenters to extend our model to the multi-source case in the unsupervised DA scenario. In our experiments on both synthetic and real-world data, JCPOT method outperforms current state-of-the-art methods and provides a computationally attractive and reliable estimation of proportions in the unlabeled target sample. In the future, we plan to extend JCPOT to estimate proportions in deep learning-based DA methods suited to for larger datasets. 

{\bf Acknowledgements.} This work was partly funded through the projects OATMIL ANR-17-CE23- 0012 and LIVES ANR-15-CE23-0026 of the French National Research Agency (ANR).

\bibliography{_refs}
\bibliographystyle{authordate1}
\end{document}